\documentclass{article}

\usepackage[preprint]{neurips_2026}

\usepackage[utf8]{inputenc} % allow utf-8 input
\usepackage[T1]{fontenc}    % use 8-bit T1 fonts
\usepackage{hyperref}       % hyperlinks
\usepackage{url}            % simple URL typesetting
\usepackage{booktabs}       % professional-quality tables
\usepackage{amsfonts}       % blackboard math symbols
\usepackage{nicefrac}       % compact symbols for 1/2, etc.
\usepackage{microtype}      % microtypography
\usepackage{xcolor}         % colors
\usepackage{xspace}
\usepackage{graphicx}
\usepackage{amsmath}

\usepackage{multicol}
\usepackage{multirow}
\usepackage[inline]{enumitem}
\usepackage{diagbox}
\usepackage{xcolor}
\usepackage[table]{xcolor}
\usepackage{cleveref}
\usepackage{adjustbox}
\usepackage{url}
\definecolor{lightred}{RGB}{255, 245, 245}
\definecolor{lightorange}{RGB}{255, 250, 240}
\definecolor{lightyellow}{RGB}{255, 255, 240}
\usepackage{siunitx}
\usepackage{tabularray}
\AddToHook{cmd/appendix/before}{%
    \crefalias{section}{appendix}%
    \crefalias{subsection}{appendix}
}
\crefname{section}{Sec.}{Secs.}
\crefname{subsection}{Sec.}{Secs.}
\crefname{table}{Tab.}{Tabs.}
\crefname{figure}{Fig.}{Figs.}
\Crefname{table}{Table}{Tables}
\Crefname{figure}{Figure}{Figures}
\Crefname{section}{Section}{Sections}

\title{Does Engram Do Memory Retrieval in Autoregressive Image Generation?}

\author{%
    Jinghao Wang$^1$ \quad Qiyuan He$^2$ \quad Chunbin Gu$^1$\thanks{Corresponding author.} \quad Pheng-Ann Heng$^1$\\
    $^1$The Chinese University of Hong Kong \quad $^2$National University of Singapore \\
    \texttt{jinghao.w@link.cuhk.edu.hk, gchb4science@gmail.com}
}

\date{}

\begin{document}
\maketitle
\begin{figure}[ht]
    \centering
    \includegraphics[width=\textwidth]{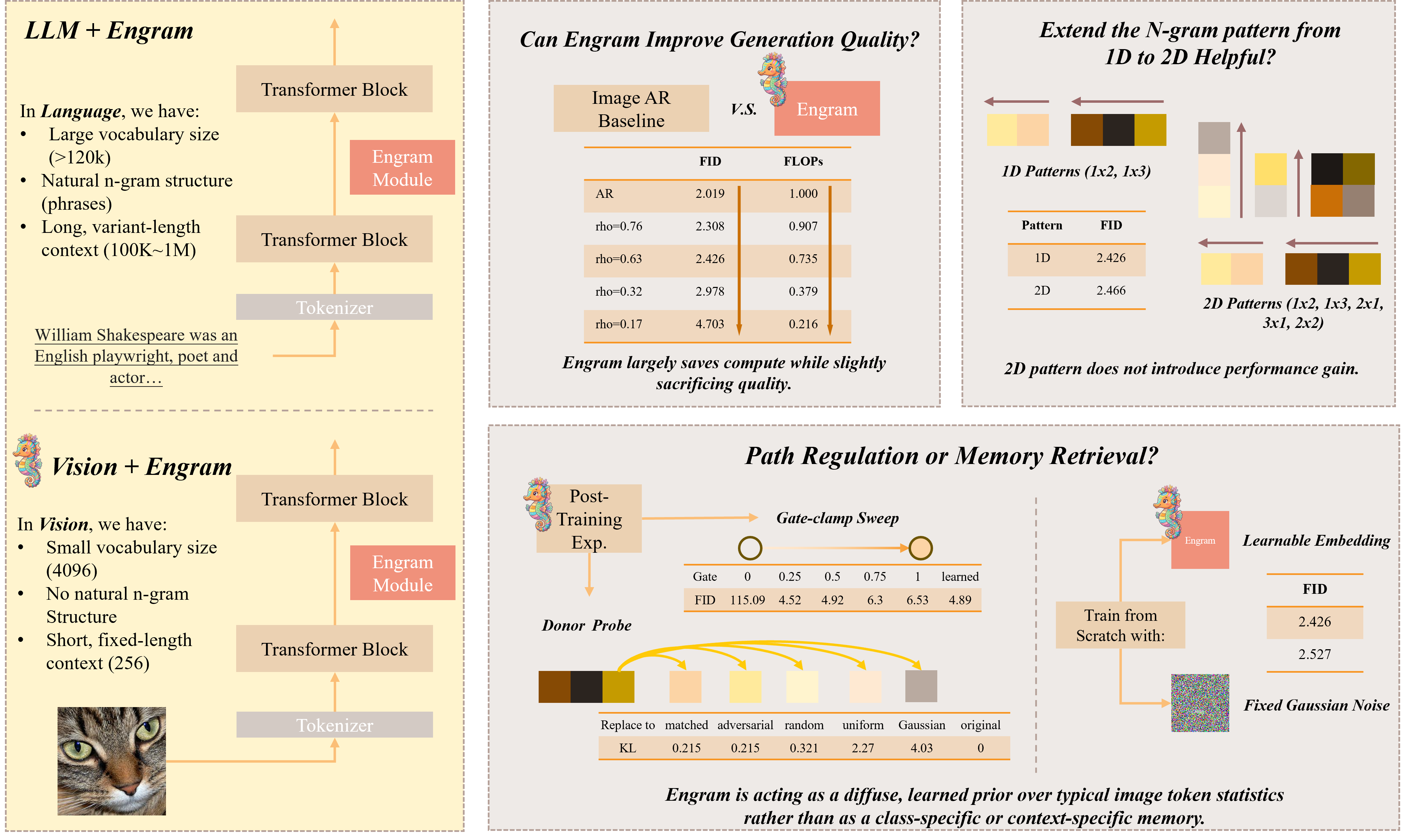}
    \caption{Left: We compare the Engram used in LLM and vision autoregressive generation. Right: we conduct three sets of experiments to understand the behavior of Engram in vision generation models.}
    \label{fig:teaser}
\end{figure}
\begin{abstract}
    The Engram module---a hash-keyed, $O(1)$ associative memory injected into
    Transformer layers---was recently shown to improve large language model
    pretraining, with the appealing interpretation that it provides a
    content-addressed shortcut to recurring local token patterns.  We ask
    whether this interpretation transfers to autoregressive (AR) image
    generation, or whether the observed gains, if any, come from a different
    mechanism.  We adapt the Engram module to vision with 2D spatial
    $n$-gram hashing, gated fusion, and KV-cache-compatible incremental
    inference, and inject it into a class-conditional AR generator trained on
    ImageNet $256{\times}256$.  Across a sweep of backbone-to-memory budget
    ratios $\rho{\in}[0.17, 0.90]$, every Engram-augmented variant trails the
    pure AR baseline in FID, indicating that the module saves backbone FLOPs
    but does not, by itself, improve sample quality.  We then probe \emph{how}
    the module is used.  A gate-clamp sweep shows that disabling the Engram
    pathway entirely is catastrophic, yet a tiny constant gate
    ($g{=}0.10$) matches or beats the learned gate---inconsistent with a
    heavily content-addressed recall mechanism.  A donor-probe experiment
    shows that swapping the hash inputs for matched, adversarial, or random
    same-class exemplars produces statistically indistinguishable next-token
    distributions, while collapsing or randomising the table degrades them by
    two to three orders of magnitude.  Finally, training a model from scratch
    with the entire memory table frozen to $\mathcal{N}(0, 1)$ noise costs
    only $\Delta\text{FID}{=}0.10$ and actually \emph{raises} Inception Score.
    Together, these findings indicate that the Engram in AR image generation
    behaves not as a content-addressed retriever but as a \emph{gated
    architectural side-pathway}: a hash-keyed residual stream whose benefit
    is dominated by the pathway itself, with the learned table contributing
    only a small distributional refinement.
\end{abstract}
\section{Introduction}\label{sec:intro}

The Engram module~\cite{cheng2026conditional}---a conditional memory module
that augments Transformer layers with a hash-based lookup table keyed by
recent token $n$-grams---was recently proposed in natural language
processing.  By hashing the local context into a large embedding table and
gated-fusing the retrieved value into the hidden state, Engram is presented
as a fast, gradient-free shortcut to recurring local patterns in language
(common phrases, idioms, collocations) and as a new axis of sparsity that
complements the dense attention and feed-forward layers of the Transformer.
The mental model behind this design is appealing: a parametric memory table
that is \emph{addressed} by the surface form of the recent context and
\emph{retrieves} a content-specific correction to the next-token logits.

Autoregressive (AR) image generation provides a sharp test for that mental
model.  By mapping images to sequences of discrete tokens via vector-quantised
encoders~\cite{van2017neural, esser2021taming} and predicting them
left-to-right with a causal Transformer, AR image
models~\cite{ramesh2021zero, yu2022scaling} produce token streams whose
local structure is, in principle, even more redundant than natural
language: with a $16{\times}16$ token grid, the same local patches---sky,
fur, brick---reappear across hundreds of classes.  If Engram is a
content-addressed retriever, it should be \emph{especially} useful here.

In this paper we adapt the Engram module to vision, plug it into a strong
class-conditional AR generator, and ask two questions:
\textbf{(Q1)} Does the resulting model improve sample quality over a pure AR
baseline at matched parameter count?
\textbf{(Q2)} If it helps (or hurts), \emph{which mechanism is responsible}:
content-addressed retrieval from the memory table, the architectural
side-pathway induced by gated fusion, or simple regularisation?

Our empirical findings, however, are negative for the memory retrieval interpretation and only mildly positive for the architecture:

\begin{enumerate}
    \item \textbf{No FID improvement at matched parameters.}  We sweep the
    backbone-to-total parameter ratio $\rho{\in}\{0.17, \dots, 0.90\}$,
    trading backbone width for memory-table capacity.  Every Engram-augmented
    variant has worse FID than the pure AR baseline ($2.02$).  The module
    saves backbone FLOPs but does not, on its own, buy a quality gain.
    Adding richer 2D $n$-gram banks ($1{\times}3$, $3{\times}1$,
    $2{\times}2$) on top of the 1D variant changes FID by less than $0.01$.
    \item \textbf{Gate clamping reveals over-mixing of memory.}  At
    inference, replacing the learned gate with a constant
    $g{\in}\{0.10, 0.25\}$ \emph{matches or beats} the learned gate, while
    $g{=}1$ degrades FID-10K from $4.89$ to $6.53$.  Disabling the pathway
    entirely ($g{=}0$) collapses generation (FID-10K $115$).  A genuine
    retriever should reward putting more weight on its outputs; this
    pathway prefers to be barely used.
    \item \textbf{The hash address does not select content.}  In a donor
    probe that swaps the tokens fed to the hash for \emph{matched},
    \emph{adversarial}, or \emph{random} same-class exemplars,
    next-token distributions are statistically indistinguishable across the
    three conditions, while collapsing the bucket index or replacing the
    table with $\mathcal{N}(0, 1)$ noise degrades them by two to three orders
    of magnitude.  The model is sensitive to whether the table is diverse
    and learned, but not to whether the queried entry semantically matches
    the context.
    \item \textbf{Most of the benefit is architectural, not content-based.}
    Training the model \emph{from scratch} with the entire memory table
    frozen to $\mathcal{N}(0, 1)$ noise costs only
    $\Delta\text{FID}{=}0.10$ relative to a fully learned table, and in
    fact \emph{raises} Inception Score from $259$ to $285$.  The bulk of
    the gain comes from the gated, hashed side-pathway---additional
    parameters and a parallel residual stream---rather than from the
    contents of the table.
\end{enumerate}

We summarise our contributions as follows:
\begin{enumerate}
    \item We adapt the Engram conditional-memory mechanism from 1D language
    to 2D vision, with spatial $n$-gram hashing over a causal
    $2{\times}2$ grid, gated fusion, per-channel LayerScale, and
    KV-cache-compatible incremental inference suitable for AR sampling.
    \item We introduce a diagnostic toolkit---\emph{gate-clamp analysis},
    a \emph{donor probe} over matched/adversarial/random/uniform/randomised
    conditions, and a \emph{from-scratch frozen-noise} training control---for
    disentangling content-addressed retrieval from architectural pathway
    effects in hash-keyed memory modules.
    \item Through a controlled sweep on class-conditional ImageNet
    $256{\times}256$, we provide evidence that, in the AR image setting,
    Engram functions as a gated architectural side-pathway rather than a
    content-addressed retriever, and that the learned memory table
    contributes only a small refinement on top of the pathway itself.
\end{enumerate}
\section{Related Work}\label{sec:related}

\paragraph{Autoregressive Image Generation.} Early work on AR image generation operated directly on pixels~\cite{vdoord2016pixel, salimans2017pixelcnn} or coarse grids~\cite{reed2017parallel}.
The introduction of VQ-VAE~\cite{van2017neural} and VQGAN~\cite{esser2021taming} enabled Transformer-based generation in a compressed discrete token space.
Subsequent work explored larger vocabularies~\cite{yu2022scaling}, improved tokenisers~\cite{yu2024titok, weber2024maskbit}, and training strategies such as random-order generation~\cite{tian2024rar} and masked prediction~\cite{chang2022maskgit}.
Our work builds on the raster-scan AR Transformer and is complementary to these advances: the Engram module can be added to any AR backbone regardless of the tokeniser or generation order.

\paragraph{Memory-Augmented Neural Networks.} External memory has a long history in neural networks, from Neural Turing Machines~\cite{graves2014ntm} and Memory Networks~\cite{weston2015memnet} to the Memorising Transformer~\cite{wu2022memorizing}, which attends over a key--value cache of past activations. These approaches typically use learned attention to route queries to memory, incurring $O(N)$ cost in memory size. In contrast, Engram~\cite{cheng2026conditional} uses deterministic hashing for $O(1)$ lookup, with the gated fusion mechanism learning \emph{whether} to use the retrieved value rather than \emph{how} to retrieve it. We inherit this efficiency and extend it to 2D spatial structure.
\section{Method}
\label{sec:method}

\subsection{Preliminary}
\label{sec:preliminary}

\textbf{Autoregressive Image Generation.} Autoregressive (AR) image generation first encodes an image into a sequence of discrete tokens using a vector-quantised (VQ) encoder such as VQ-VAE~\cite{van2017neural} or VQGAN~\cite{esser2021taming}, then models the resulting sequence with a causal Transformer.
A typical encoder partitions the image into a regular spatial grid of patches (e.g.\ $16{\times}16 = 256$ patches for a $256{\times}256$ image) and maps each patch to a discrete index drawn from a finite codebook via nearest-neighbour lookup in a learned embedding space.
Flattening this grid in raster-scan order yields a token sequence $\mathbf{z} = (z_1, z_2, \ldots, z_T)$, where each $z_t \in \{1, \ldots, V\}$ and $V$ is the codebook size.
The model factorises the joint distribution autoregressively:
\begin{equation}\label{eq:ar}
    p(\mathbf{z}) = \prod_{t=1}^{T} p(z_t \mid z_{<t}),
\end{equation}
where each conditional is parameterised by a causal Transformer trained with cross-entropy loss.

\textbf{Engram.} The Engram module~\cite{cheng2026conditional} augments a Transformer layer with a hash-based external memory that enables $O(1)$-cost retrieval of static local patterns.
Standard Transformers lack a native knowledge-lookup primitive and must simulate retrieval through sequential layers of attention and feed-forward computation~\cite{cheng2026conditional}; Engram addresses this by offloading static pattern recall to an explicit embedding table. The core components are as follows:
\begin{itemize}[topsep=0pt, leftmargin=*]
    \item \textbf{Multi-head, multi-order hashing.}
Rather than maintaining a single table, Engram instantiates separate embedding tables for each n-gram order $n \in \{2, \ldots, N\}$ and each of $K$ independent hash heads, yielding tables $\{\mathbf{E}_{n,k} \in \mathbb{R}^{M_{n,k} \times d_{\rm head}}\}$ of prime size $M_{n,k}$.
For position $t$, the $n$-gram context $g_{t,n} = (z_{t-n+1}, \ldots, z_t)$ is mapped by a lightweight multiplicative-XOR hash $\varphi_{n,k}$ to slot index $\varphi_{n,k}(g_{t,n})$, and the retrieved vectors are concatenated across all orders and heads to form the aggregated memory vector:
\begin{equation}\label{eq:engram-retrieve}
    \mathbf{e}_t \;=\; \bigl\|_{n=2}^{N}\; \bigl\|_{k=1}^{K}\; \mathbf{E}_{n,k}\bigl[\varphi_{n,k}(g_{t,n})\bigr]
    \;\in\; \mathbb{R}^{d_{\rm mem}},
\end{equation}
where $d_{\rm mem} = (N-1) \cdot K \cdot d_{\rm head}$.
Using $K$ independent hash functions per order substantially reduces hash-collision noise relative to a single-head design.
    \item \textbf{Context-aware gating and fusion.} The retrieved vector $\mathbf{e}_t$ is context-independent and may contain noise from collisions or polysemy.
Engram resolves this ambiguity via learnable key and value projections $\mathbf{W}_K, \mathbf{W}_V$ applied to $\mathbf{e}_t$:
\begin{equation}\label{eq:engram-gate}
    g_t = \sigma\!\left(\frac{\mathrm{RMSNorm}(\mathbf{h}_t)^{\!\top}\,\mathrm{RMSNorm}(\mathbf{W}_K \mathbf{e}_t)}{\sqrt{d}}\right),
    \qquad
    \tilde{\mathbf{v}}_t = g_t \cdot \mathbf{W}_V \mathbf{e}_t.
\end{equation}
The current hidden state $\mathbf{h}_t$—which has aggregated global context via preceding attention—acts as a dynamic query: if the retrieved memory contradicts the context, $g_t$ tends to zero, suppressing retrieval noise.
$\mathbf{W}_V$ is implemented as a two-layer MLP when $d_{\rm mem} > d$ (i.e.\ when the memory dimension exceeds the backbone hidden size).
Finally, a lightweight depthwise causal convolution (kernel size 4, SiLU activation, zero-initialised) is applied to the sequence of gated values $\tilde{\mathbf{V}}$ to expand the local receptive field, yielding the fused output:
\begin{equation}\label{eq:engram}
    \mathbf{h}_t \leftarrow \mathbf{h}_t + g_t \cdot \mathbf{v}_{h_t},
\end{equation}
where $\mathbf{v}_{h_t} \equiv (\mathbf{W}_V \mathbf{e}_t + \mathrm{SiLU}(\mathrm{Conv1D}(\mathrm{RMSNorm}(\tilde{\mathbf{V}})))_t)$ subsumes the convolution refinement.
Because hashing is deterministic and computed solely from the raw token IDs, the lookup is $O(1)$ in memory size; gradients flow only through $\mathbf{E}_{n,k}$ and $g_t$ (not through the hash functions themselves).
\end{itemize}

\subsection{Confirmation of Existance of N-gram Structure in Visual Tokens}
\label{sec:ngram-in-vision}
Since the image AR models are trained not on the pure pixel space but in the token space provided by the image tokenizer and unlike natural language, where $n$-grams (contiguous word sequences) are well-established, it is unclear whether discrete image tokens possess analogous structure, or whether this structure depends on the tokeniser. Hence, a prerequisite for adapting Engram to vision is that \emph{visual token sequences must exhibit recurrent local $n$-gram patterns}.Otherwise, the hash-based lookup has nothing to retrieve. 

\textbf{DINO-stratified Jaccard analysis.} We measure the $n$-gram overlap between pairs of images as a function of their semantic similarity. For a patch shape $(p_r{\times}p_c)$ and two token grids $A$ and $B$, the 2D Jaccard similarity is
\begin{equation}\label{eq:jaccard}
    J_{p_r,p_c}(A, B) \;=\;
    \frac{\bigl|\mathcal{N}_{p_r,p_c}(A) \cap \mathcal{N}_{p_r,p_c}(B)\bigr|}
         {\bigl|\mathcal{N}_{p_r,p_c}(A) \cup \mathcal{N}_{p_r,p_c}(B)\bigr|},
\end{equation}
where $\mathcal{N}_{p_r,p_c}(\cdot)$ is the multiset of all $(p_r{\times}p_c)$
patches extracted from the grid.  We evaluate five shapes:
$1{\times}2$, $2{\times}1$, $1{\times}3$, $3{\times}1$, and $2{\times}2$.
Pairs are grouped into bins of DINOv2~\cite{oquab2023dinov2} cosine similarity
(range $[0.50, 1.00]$, 10 bins of width 0.05, up to 500 pairs per bin).
All pairs are drawn from a uniform random subsample of 15\,000 training images
(seed 42); \emph{no curated near-duplicate index is used}.
A ``random'' baseline computes $J$ over pairs drawn without regard to similarity.

\begin{figure}[ht]
    \centering
    \includegraphics[width=\textwidth]{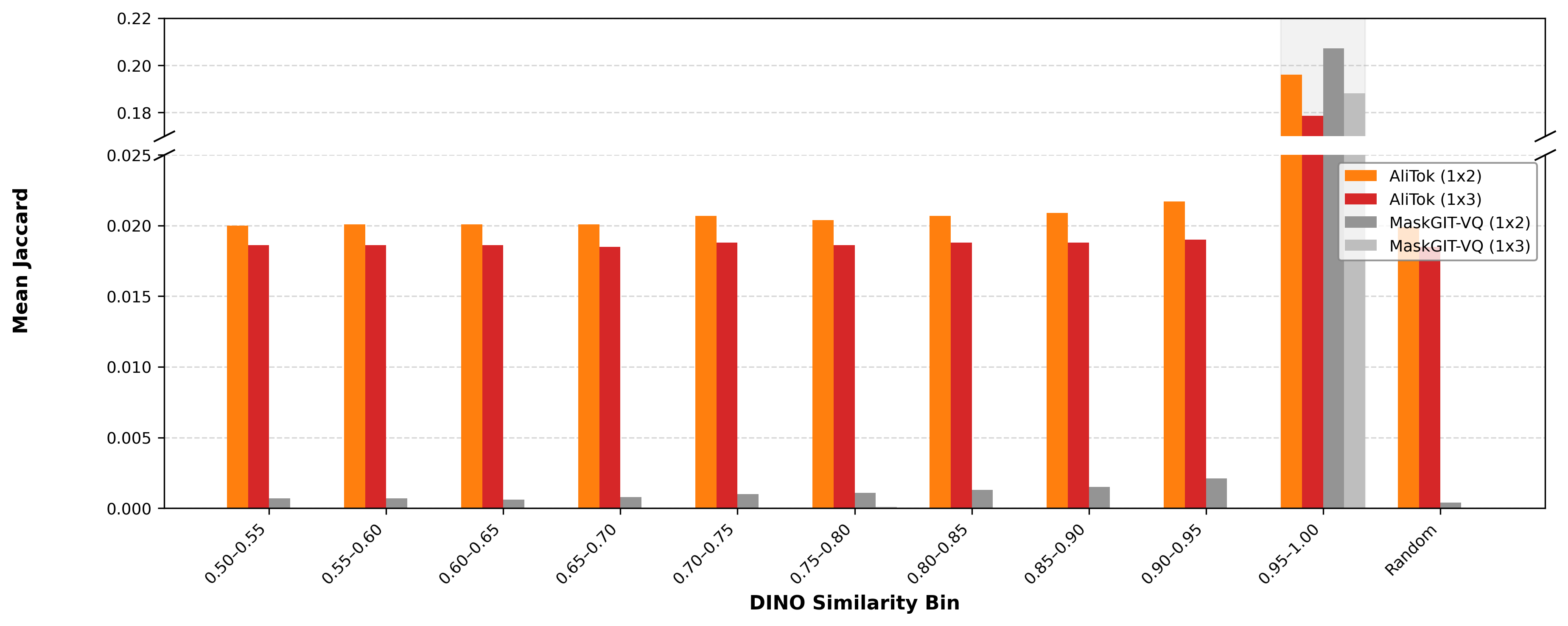}
    \caption{\textbf{N-gram Jaccard similarity stratified by DINO similarity.}  AliTok tokens exhibit consistent local patch patterns that grow stronger with semantic similarity, while MaskGIT-VQ tokens do not.}
    \label{fig:ngram}
\end{figure}

\textbf{Result.} \Cref{fig:ngram} shows the mean Jaccard similarity as a grouped bar
chart for each shape, stratified by DINO similarity tier, for two tokenisers:
\textbf{AliTok}~\cite{wu2025towards} (codebook size 4096, 256 semantic tokens
arranged as a $16{\times}16$ grid; decoder uses \emph{causal} attention) and
\textbf{MaskGIT-VQ}~\cite{chang2022maskgit} (codebook size 1024, 256 tokens
on the same $16{\times}16$ grid; decoder uses bidirectional attention).

AliTok produces tokens with consistent local patterns at all levels of similarity. MaskGIT-VQ, by contrast, collapses to near-zero for all shapes at low similarity and reaches meaningful overlap only at the extreme $\geq$0.95 tier
where images are near-duplicates.

This result motivates our choice of \textbf{AliTok as the tokeniser} for all
subsequent experiments.  An Engram module that hashes spatial-patch contexts
can only be effective when the token space exhibits recurrent local
patterns---a condition AliTok satisfies but MaskGIT-VQ does not.

%--------------------------------------------------------------------------
\subsection{Engram for Autoregressive Image Generation}\label{sec:vengram}
%--------------------------------------------------------------------------

\textbf{Base model.} We choose our base autoregressive model as (\textbf{AR-B}): a 24-layer causal Transformer with hidden size~768, 16 attention heads, 2D Rotary Position Embeddings (RoPE), SwiGLU feed-forward layers, RMSNorm, and QK-normalisation with RMSNorm. The model totals approximately 177M parameters.

\textbf{Vision adaptation.} We list the differences from the NLP Engram as follows:
\begin{enumerate}[nosep,leftmargin=*]
    \item \textbf{Context n-gram definition.}  At each grid position $(i,j)$ in the $16{\times}16$ raster order, we collect a set of causal spatial neighbours and hash them with $H$ independent hash functions into indices $\{h_1, \ldots, h_H\}$ within a table of $V_{\text{mem}}$ entries.We maintain two independent banks: a 2-gram bank $\{t, t{-}1\}$ and a 3-gram bank $\{t, t{-}1, t{-}2\}$.\
    \item \textbf{No tokenizer compression.}  Unlike the NLP setting where the Engram operates with a compressed tokenizer, we keep the original Alitok tokenized results unchanged.
\end{enumerate}

By default we place engram modules at layers $\{0, 6, 12, 18\}$ with
$H{=}4$ hash heads, $V_{\text{mem}}{=}36\,715$ memory slots,
and head dimension $d_h{=}64$. In addition, Alitok introduces a set of 256 images tokens, plus 16 auxiliary tokens and 1 class tokens. We only apply the Engram to the 256 image tokens.
\section{Experiments \& Analysis}
\label{sec:experiments}
% define a counter using Roman number
\newcounter{example}
\stepcounter{example}
\subsection{Experimental Setup}
\label{sec:setup}

\noindent\textbf{Dataset and Tokeniser.} We train on the ImageNet~\cite{deng2009imagenet} training set (1.28M images, 1000 classes) at $256{\times}256$ resolution. Based on the analysis above, we use the AliTok~\cite{wu2025towards} tokeniser, which maps each image to 256 semantic tokens (plus 17 aux tokens)from a codebook of size 4096.

\noindent\textbf{Training and Evaluation.} All models are trained with AdamW (lr $4{\times}10^{-4}$, cosine schedule, $\beta_1{=}0.9$, $\beta_2{=}0.96$, weight decay $0.03$) and bfloat16 mixed precision. Effective batch size is 2048 across 8 GPUs.  Each condition in the four-way ablation is trained for 125K steps under identical settings (200 epochs). We report FID-50K~\cite{heusel2017gans} and Inception Score (IS) computed with classifier-free guidance (cosine-power schedule, $\text{cfg}_{\max}{=}16.0$, $\alpha{=}1.8$).

\subsection{Finding \Roman{example}: The Engram-Augmented AR Saves FLOPs But Slightly Sacrifices FID}
\label{sec:best}
\stepcounter{example}
\textbf{Setup.} As we are working one a dense transformer model, we control the the \textbf{backbone ratio} $\rho = \text{backbone params} / \text{total params}$; $\rho{=}1$ is the pure AR baseline and lower $\rho$ allocates more budget to the engram memory tables. Since, we have to keep the depth of the model constant so that the Engram module can be injected at the same layer, we adjust the width of the backbone parameters by varying the size of hidden dimension, range from 320 to 768. The head dimension of the Engram module is fixed at 64, and the number of hash heads is fixed at 4. The size of the memory table is varied from 52831 to 5243. It yields a family of Engram-augmented variants with $\rho = \{0.17, 0.32, 0.41, 0.51, 0.63, 0.76, 0.90\}$ where the lowest $\rho$ corresponds to the largest memory table and the smallest backbone, and vice versa for the highest $\rho$. $\rho=1$ stands for the pure AR baseline without any Engram module.

\begin{figure}[t]
    \centering
    \includegraphics[width=1\textwidth]{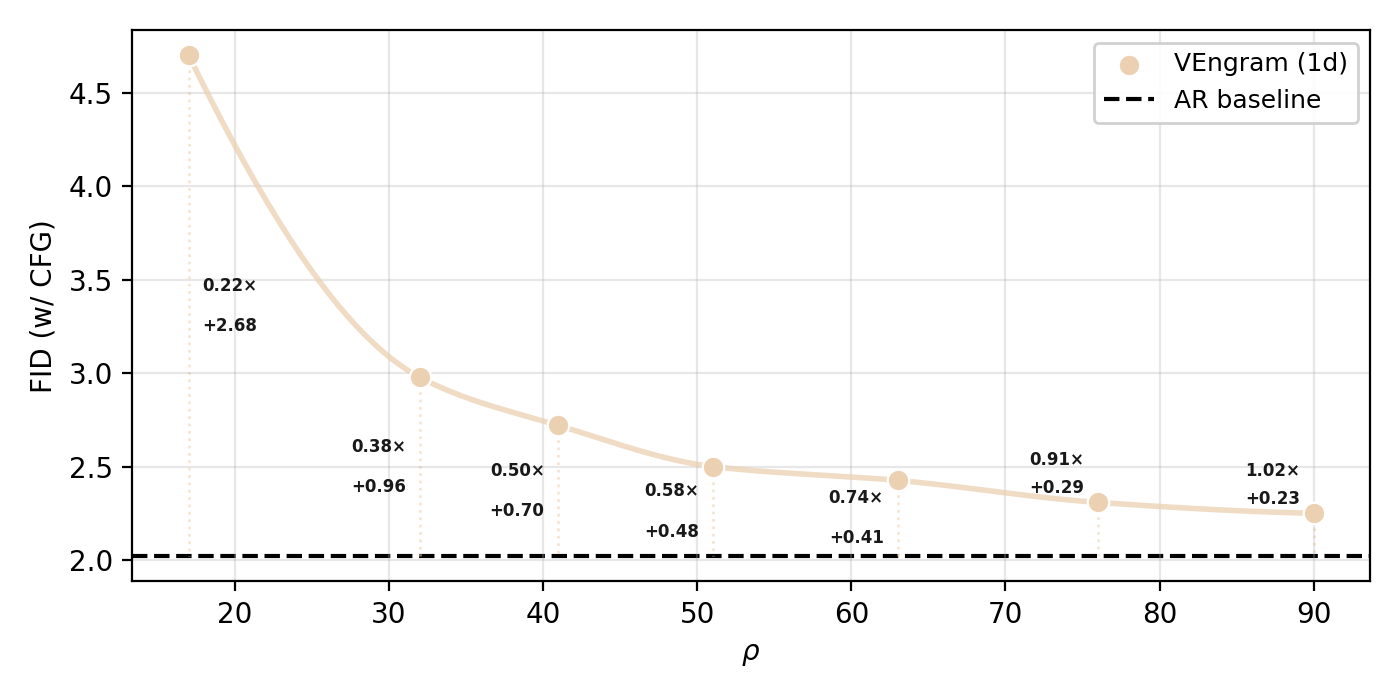}
    \caption{\textbf{Proportion sweep.} The Engram-augmented AR saves FLOPs by offloading local-pattern reconstruction to the Engram memory tables, but this comes with a slight FID cost.}
    \label{fig:rho-sweep}
\end{figure}

\textbf{Result.} In \Cref{fig:rho-sweep}, we observe that all Engram-augmented variants with $\rho$ from 17\% to 90\% have worse FID than the AR baseline (2.02). This is a promising result given that Engram is a lightweight module designed for efficiency, but it also motivates our subsequent analysis to understand why the Engram module does not produce a stronger improvement in FID.

\subsection{Finding \Roman{example}: 2D N-gram Dose Not Improve FID}
\label{sec:variants}
\stepcounter{example}

\textbf{Setup.} In contract to language lived in 1D space, image is 2D. It replaces the sequential offsets with the combination of the 2D spatial neighbours:
\begin{enumerate*}
    \item Bank 0 ($1{\times}2$): $\{z_t,\; z_{\text{left}}\}$
    \item Bank 1 ($2{\times}1$): $\{z_t,\; z_{\text{top}}\}$
    \item Bank 2 ($1{\times}3$): $\{z_t,\; z_{\text{left}},\; z_{\text{left left}}\}$
    \item Bank 3 ($3{\times}1$): $\{z_t,\; z_{\text{top}},\; z_{\text{top top}}\}$
    \item Bank 4 ($2{\times}2$): $\{z_t,\; z_{\text{left}},\; z_{\text{top}},\; z_{\text{diag}}\}$
\end{enumerate*}.
The 2D variant is designed to capture both horizontal and vertical token co-occurrences simultaneously.

\textbf{Result.} The 2D variant achieves an FID of $4.24$, which is nearly identical to the 1D baseline ($4.23$).  This suggests that the additional vertical and diagonal context captured by the 2D n-grams does not translate into a measurable improvement in generation quality.

\subsection{Finding \Roman{example}: The Engram Uses Memory as a Learned Distributional Prior, Not as Content Retrieval}
\label{sec:ablation-design}
\stepcounter{example}

\noindent\textbf{Gate-Clamp Evaluation.} To probe whether the model uses memory selectively, we evaluate the trained \textsc{Real} model ($\rho{=}0.63$, $h{=}640$, 1D) with the gate globally clamped to fixed values $g\in\{0.00, 0.10, 0.25, 0.50, 0.75, 1.00\}$ as well as with the learned (token-dependent) free gate.  All conditions use FID-10K with identical CFG settings.

\begin{table}
    \centering
    \caption{\textbf{Gate-clamp sweep} on the final \textsc{Real} checkpoint
    (FID-10K diagnostic; the headline FID-50K of this checkpoint is $2.43$, so
    absolute numbers below carry the standard 10K finite-sample bias and are
    only comparable \emph{within} this table).  Disabling the engram pathway
    ($g{=}0$) collapses generation, but a small constant gate is sufficient to
    recover---and even slightly outperform---the learned gate.}
    \label{tab:gate-clamp}
    \begin{tabular}{lc}
    \toprule
    Gate value & FID-10K$\downarrow$ \\
    \midrule
    0.00 (no memory) & 115.09 \\
    0.10 & \textbf{4.49} \\
    0.25 & 4.52 \\
    Learned (free gate) & 4.89 \\
    0.50 & 4.92 \\
    0.75 & 6.30 \\
    1.00 (max memory) & 6.53 \\
    \bottomrule
    \end{tabular}
\end{table}

\Cref{tab:gate-clamp} shows two things.  First, the engram pathway is
\emph{necessary}: clamping $g{=}0$ removes any memory contribution and
catastrophically degrades FID (4.89 $\to$ 115).  Second, the learned gate is
slightly \emph{over-mixing}: a constant clamp at $g{\in}\{0.10, 0.25\}$ already
matches or improves on the free gate, while pushing the gate higher
monotonically degrades quality.  Together this indicates that the engram
contribution should be small but non-zero---consistent with a low-rank
correction to the AR backbone rather than a heavily content-addressed
recall mechanism.

\noindent\textbf{Donor Probe.} If Engram functions as a nearest-neighbour retrieval module, then replacing
its memory content with \emph{matched} exemplars (same-class generations) should
produce noticeably better predictions than \emph{adversarial} (different-class)
or \emph{random} exemplars.  We test this with a donor probe that, while
teacher-forcing the model on $N{=}2{,}000$ self-generated reference sequences,
swaps the tokens fed into the engram hash for one of six donor conditions and
compares the resulting next-token logits to those produced with the model's
own context.  We measure the mean rank of the real top-1 token in the donor
distribution, the KL from the real distribution, and the top-5 overlap.

\begin{table}
    \centering
    \caption{\textbf{Donor probe results} ($N{=}2{,}000$ sequences, 4096
    positions each).  \textsc{Matched}, \textsc{Adversarial} and \textsc{Random}
    donors all produce nearly identical perturbations to the next-token
    distribution, while \textsc{Uniform} (collapse all hash buckets to bucket
    0) and \textsc{Randomized} (replace memory table with $\mathcal{N}(0,1)$
    noise) cause two-to-three orders of magnitude larger degradations.  This
    weakens a strict retrieval interpretation: the model uses the engram for
    its \emph{distributional structure}, not for class-conditional content
    addressing.}
    \label{tab:donor}
    \begin{tabular}{lccc}
    \toprule
    Donor condition & Mean rank of real top-1$\downarrow$ & KL from real$\downarrow$
        & Top-5 overlap$\uparrow$ \\
    \midrule
    Matched      & 4.46    & 0.215 & 0.634 \\
    Adversarial  & 4.62    & 0.216 & 0.634 \\
    Random       & 6.07    & 0.321 & 0.613 \\
    \midrule
    Uniform      & 292.94  & 2.27  & 0.189 \\
    Randomized   & \textbf{1313.50} & \textbf{4.03} & \textbf{0.014} \\
    \bottomrule
    \end{tabular}
\end{table}

\Cref{tab:donor} shows that \textsc{Matched}, \textsc{Adversarial}, and
\textsc{Random} donors produce nearly identical output distributions
($\Delta\text{rank}{<}2$, $\Delta\text{KL}{<}0.11$, top-5 overlap differs by
$<0.03$).  In particular, \emph{matched} is statistically indistinguishable
from \emph{adversarial}, indicating that the engram is not encoding
class-specific content.  In contrast, \textsc{Uniform} (collapsing all address
diversity to a single bucket) inflates the rank by $\sim$60$\times$ and
\textsc{Randomized} (destroying the learned table while keeping the keying
pathway) by nearly $\sim$300$\times$.  Hence the engram is sensitive to
whether memory remains \emph{diverse and learned}, but it does \textbf{not}
preferentially select semantically matched content---undermining a strict
retrieval interpretation and supporting a view of the engram as a learned
distributional prior over local image-token statistics.

\subsection{Finding \Roman{example}: Most of the Engram's Benefit Comes From the Architectural Pathway, Not From Memory Content}
\stepcounter{example}

The donor probe in \Cref{sec:ablation-design} establishes that, \emph{at
inference time}, the engram is sensitive to whether memory remains diverse
and learned, but does not select semantically matched content.  This still
leaves open whether, when training \emph{from scratch}, a meaningful learned
memory table is required at all.  We test this by training a fourth
condition---\textsc{Randomized}---in which the memory embedding table is
replaced with frozen $\mathcal{N}(0, 1)$ noise (not updated during training)
while every other component (hashing, gating, AR backbone) is unchanged.
This preserves the engram \emph{pathway} but destroys any learnable
content.

\begin{table}[t]
    \centering
    \caption{\textbf{Real vs.\ randomized engram, trained from scratch under
    identical settings.}  FID-50K and IS reported at step 125K with full CFG
    ($\text{cfg}_{\max}{=}16$, $\alpha{=}1.8$).  Replacing the entire memory
    table with frozen Gaussian noise costs only $\Delta\text{FID}{=}0.10$
    and actually \emph{increases} IS, indicating that the learned memory
    contributes a small refinement on top of the architectural pathway,
    while the noise variant produces sharper but slightly less faithful
    samples.}
    \label{tab:real-vs-rand-trained}
    \begin{tabular}{lcc}
    \toprule
    Condition & FID-50K$\downarrow$ & IS$\uparrow$ \\
    \midrule
    \textsc{Real} (learned memory)        & \textbf{2.426} & 259.42 \\
    \textsc{Randomized} (frozen $\mathcal{N}(0, 1)$ memory) & 2.527 & \textbf{284.70} \\
    \bottomrule
    \end{tabular}
\end{table}

\Cref{tab:real-vs-rand-trained} reveals that The architectural pathway carries most of the benefit.
Replacing every value in the memory table with frozen Gaussian noise
costs only $\Delta\text{FID}{=}0.10$ relative to a fully learned memory.
By contrast, removing the engram pathway entirely (gate clamp $g{=}0$ in
\Cref{tab:gate-clamp}) collapses FID-10K to $115.09$.  Hence the bulk of
the engram's contribution is attributable to the gated, hashed
side-pathway---additional parameters, content-conditioned gating, and a
parallel residual stream---rather than to the specific contents of the
memory table.  This is consistent with the donor-probe finding that the
engram functions as a learned distributional prior, not as a
content-addressed retriever.

\subsection{Visualization}
\label{sec:visualization}
We qualitatively compare the AR baseline and three Engram-augmented variants ($\rho{\in}\{0.17-0.90\}$) plus 2D Engram with $\rho=0.63$ on the same ten ImageNet classes generated
with the same random seed and identical CFG settings, so that any per-column
difference is attributable to the model only.

\begin{figure}[t]
    \centering
    \includegraphics[width=\textwidth]{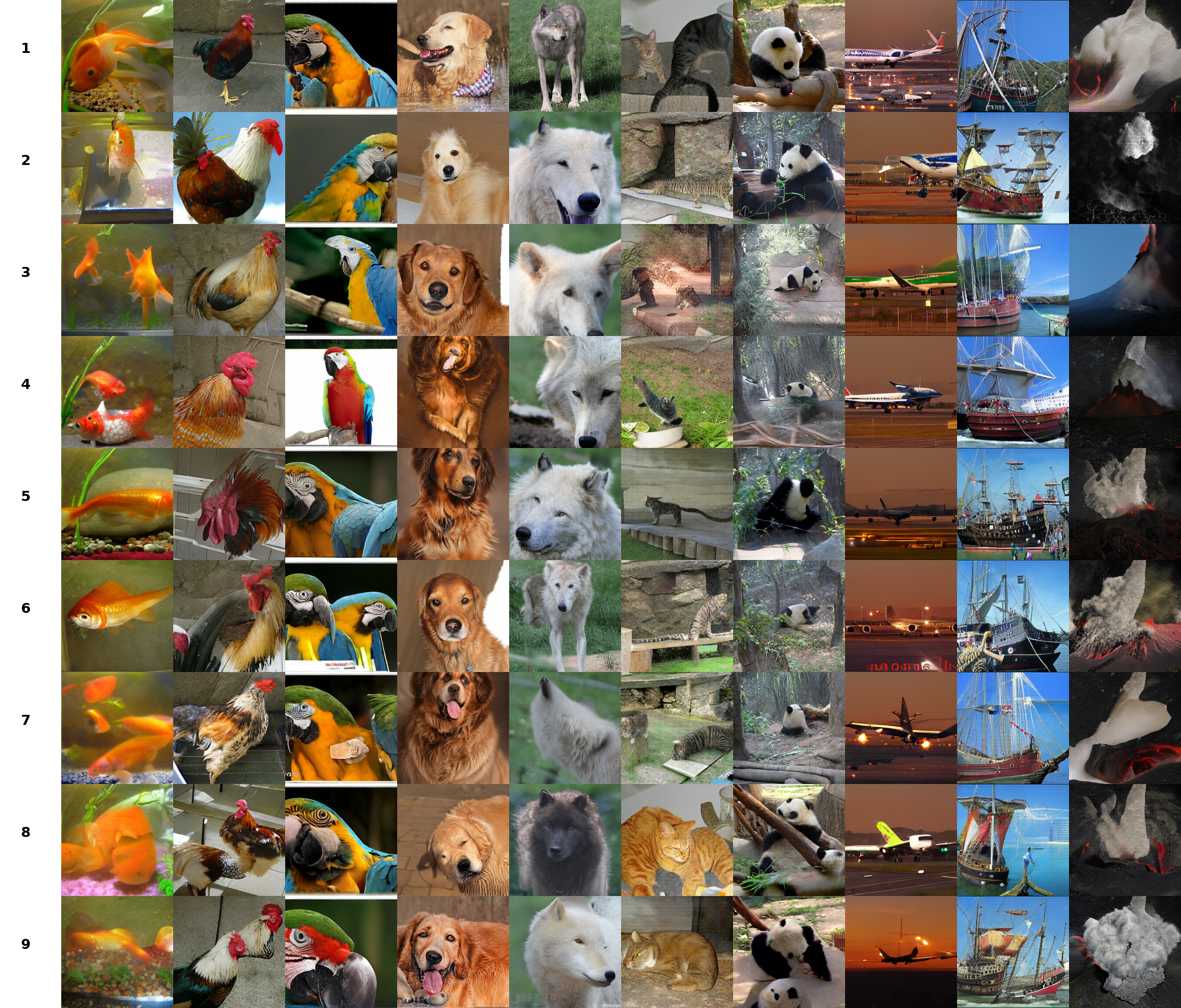}
    \caption{\textbf{Qualitative comparison.}  Rows (top to bottom):
    (1) AR baseline, (2-8) Engram-augmented variants at $\rho{\in}\{0.17-0.90\}$, and (9) 2D Engram with $\rho=0.63$ with
    row labels rendered in the left margin.
    Classes (left to right):
    goldfish, cock, macaw, golden retriever, white wolf, tiger cat, panda,
    airliner, pirate ship, volcano.}
    \label{fig:samples}
\end{figure}
\section{Conclusion}\label{sec:conclusion}

We asked whether the hash-keyed Engram memory module helps autoregressive
image generation, and---if so---through what mechanism: content-addressed
retrieval, or an architectural side-pathway with a gated residual stream. We adapted the Engram mechanism from 1D language to 2D
vision with spatial $n$-gram hashing over a causal $2{\times}2$ grid,
gated fusion, and KV-cache-compatible incremental inference.  Across a
controlled budget sweep $\rho{\in}[0.17, 0.90]$ on class-conditional
ImageNet $256{\times}256$, every Engram-augmented variant has worse FID
than a pure AR baseline at matched parameter count, and the additional 2D
$n$-gram banks do not measurably improve over the 1D variant.  The module
saves backbone FLOPs, but it does not, by itself, buy a quality gain.

We then probed \emph{how} the module is used by combining two
post-training diagnostics on a trained Engram-augmented checkpoint
($\rho{=}0.63$)---\textbf{gate-clamp analysis} and a \textbf{donor
probe}---with one \textbf{frozen-noise} training control in which the
entire memory table is replaced by frozen $\mathcal{N}(0, 1)$ noise and
the model is retrained from scratch under otherwise identical settings.
Together they give the following picture:
\begin{itemize}
    \item \textbf{The pathway is necessary, but should be barely used.}
    Disabling the gate ($g{=}0$) collapses FID-10K from $4.89$ to $115$;
    yet a constant $g{\in}\{0.10, 0.25\}$ matches or beats the learned
    gate, and $g{=}1$ monotonically degrades quality.  The model prefers
    a small, near-constant injection from the memory pathway.
    \item \textbf{The hash address is not used to select content.}  The
    donor probe shows that matched, adversarial, and random same-class
    exemplars yield statistically indistinguishable next-token
    distributions, while collapsing the bucket index or replacing the
    table with $\mathcal{N}(0, 1)$ noise degrades them by two to three
    orders of magnitude.  The model uses the table for distributional
    structure, not for class-conditional content addressing.
    \item \textbf{The table itself is largely incidental.}  In the
    frozen-noise control, training a model from scratch with the entire
    memory table frozen to $\mathcal{N}(0, 1)$ noise costs only
    $\Delta\text{FID}{=}0.10$ and \emph{raises} Inception Score from
    $259$ to $285$.  The bulk of the measurable behaviour comes from the
    gated, hashed side-pathway---an additional parameter budget and a
    parallel residual stream---rather than from the specific contents
    of the memory table.
\end{itemize}

We therefore conclude that, in autoregressive image generation, the
Engram module is best described as a \textbf{gated architectural
side-pathway}: a hash-keyed residual stream that is necessary at inference
but only weakly content-addressed, with the learned table contributing
only a small distributional refinement on top of the pathway itself.
This reading is at odds with the content-addressed retrieval interpretation
that motivated the original Engram design and suggests that, on visual
token streams produced by current tokenisers, hash-keyed memory in its
present form is better viewed as a regulariser-like architectural prior
than as a true memory.

\textbf{Limitations.}  Due to computational constrait, our study is restricted to a dense
AR generator at moderate scale; we do not train XL-size model and MoE
variants, or training regimes (e.g.\ extreme low-data, long-tail
class-conditional, or higher resolutions) where genuine memory may matter
more.
\clearpage
\bibliographystyle{plain}
\bibliography{main}

@inproceedings{chang2022maskgit,
  title={MaskGIT: Masked Generative Image Transformer},
  author={Chang, Huiwen and Zhang, Han and Jiang, Lu and Liu, Ce and Freeman, William T.},
  booktitle={{CVPR}},
  year={2022}
}

@inproceedings{deng2009imagenet,
  title={ImageNet: A Large-Scale Hierarchical Image Database},
  author={Deng, Jia and Dong, Wei and Socher, Richard and Li, Li-Jia and Li, Kai and Fei-Fei, Li},
  booktitle={{CVPR}},
  year={2009}
}

@article{oquab2023dinov2,
  title={Dinov2: Learning robust visual features without supervision},
  author={Oquab, Maxime and Darcet, Timoth{\'e}e and Moutakanni, Th{\'e}o and Vo, Huy and Szafraniec, Marc and Khalidov, Vasil and Fernandez, Pierre and Haziza, Daniel and Massa, Francisco and El-Nouby, Alaaeldin and others},
  journal={arXiv preprint arXiv:2304.07193},
  year={2023}
}

@article{cheng2026conditional,
  title={Conditional memory via scalable lookup: A new axis of sparsity for large language models},
  author={Cheng, Xin and Zeng, Wangding and Dai, Damai and Chen, Qinyu and Wang, Bingxuan and Xie, Zhenda and Huang, Kezhao and Yu, Xingkai and Hao, Zhewen and Li, Yukun and others},
  journal={arXiv preprint arXiv:2601.07372},
  year={2026}
}

@inproceedings{esser2021taming,
  title={Taming transformers for high-resolution image synthesis},
  author={Esser, Patrick and Rombach, Robin and Ommer, Bjorn},
  booktitle={{CVPR}},
  year={2021}
}

@article{graves2014ntm,
  title={Neural Turing Machines},
  author={Graves, Alex and Wayne, Greg and Danihelka, Ivo},
  journal={arXiv preprint arXiv:1410.5401},
  year={2014}
}

@article{wu2025towards,
  title={Towards Sequence Modeling Alignment between Tokenizer and Autoregressive Model},
  author={Wu, Pingyu and Zhu, Kai and Liu, Yu and Tang, Longxiang and Yang, Jian and Peng, Yansong and Zhai, Wei and Cao, Yang and Zha, Zheng-Jun},
  journal={arXiv preprint arXiv:2506.05289},
  year={2025}
}

@inproceedings{heusel2017gans,
  title={{GANs} Trained by a Two Time-Scale Update Rule Converge to a Local Nash Equilibrium},
  author={Heusel, Martin and Ramsauer, Hubert and Unterthiner, Thomas and Nessler, Bernhard and Hochreiter, Sepp},
  booktitle={Advances in Neural Information Processing Systems (NeurIPS)},
  year={2017}
}

@inproceedings{reed2017parallel,
  title={Parallel Multiscale Autoregressive Density Estimation},
  author={Reed, Scott and van den Oord, A{\"a}ron and Kalchbrenner, Nal and Colmenarejo, Sergio G{\'o}mez and Wang, Ziyu and Belov, Dan and de Freitas, Nando},
  booktitle={Proceedings of the International Conference on Machine Learning (ICML)},
  year={2017}
}

@inproceedings{salimans2017pixelcnn,
  title={{PixelCNN++}: Improving the {PixelCNN} with Discretized Logistic Mixture Likelihood and Other Modifications},
  author={Salimans, Tim and Karpathy, Andrej and Chen, Xi and Kingma, Diederik P.},
  booktitle={Proceedings of the International Conference on Learning Representations (ICLR)},
  year={2017}
}

@article{tian2024rar,
  title={Randomized Autoregressive Visual Generation},
  author={Yu, Qihang and He, Ju and Deng, Xueqing and Shen, Xiaohui and Chen, Liang-Chieh},
  journal={arXiv preprint arXiv:2411.00776},
  year={2024}
}

@article{van2017neural,
  title={Neural discrete representation learning},
  author={Van Den Oord, Aaron and Vinyals, Oriol and others},
  journal={{NeurIPS}},
  year={2017}
}

@inproceedings{vdoord2016pixel,
  title={Pixel Recurrent Neural Networks},
  author={van den Oord, Aaron and Kalchbrenner, Nal and Kavukcuoglu, Koray},
  booktitle={Proceedings of the International Conference on Machine Learning (ICML)},
  year={2016}
}

@article{weber2024maskbit,
  title={{MaskBit}: Embedding-free Image Generation via Bit Tokens},
  author={Weber, Mark and Yu, Lijun and Yu, Qihang and Deng, Xueqing and Shen, Xiaohui and Cremers, Daniel and Chen, Liang-Chieh},
  journal={Transactions on Machine Learning Research (TMLR)},
  year={2024}
}

@inproceedings{weston2015memnet,
  title={Memory Networks},
  author={Weston, Jason and Chopra, Sumit and Bordes, Antoine},
  booktitle={Proceedings of the International Conference on Learning Representations (ICLR)},
  year={2015}
}

@inproceedings{wu2022memorizing,
  title={Memorizing Transformers},
  author={Wu, Yuhuai and Rabe, Markus N. and Hutchins, DeLesley and Szegedy, Christian},
  booktitle={Proceedings of the International Conference on Learning Representations (ICLR)},
  year={2022}
}

@article{yu2022scaling,
  title={Scaling autoregressive models for content-rich text-to-image generation},
  author={Yu, Jiahui and Xu, Yuanzhong and Koh, Jing Yu and Luong, Thang and Baid, Gunjan and Wang, Zirui and Vasudevan, Vijay and Ku, Alexander and Yang, Yinfei and Ayan, Burcu Karagol and others},
  journal={arXiv preprint arXiv:2206.10789},
  year={2022}
}

@article{yu2024titok,
  title={An Image is Worth 32 Tokens for Reconstruction and Generation},
  author={Yu, Qihang and Weber, Mark and Deng, Xueqing and Shen, Xiaohui and Cremers, Daniel and Chen, Liang-Chieh},
  journal={arXiv preprint arXiv:2406.07550},
  year={2024}
}

@inproceedings{ramesh2021zero,
  title={Zero-shot text-to-image generation},
  author={Ramesh, Aditya and Pavlov, Mikhail and Goh, Gabriel and Gray, Scott and Voss, Chelsea and Radford, Alec and Chen, Mark and Sutskever, Ilya},
  booktitle={{ICML}},
  year={2021},
}
% \clearpage
% \input{checklist.tex}
\end{document}